\documentclass{article}
\usepackage{ijcai24}

\usepackage{times}
\usepackage{soul}
\usepackage{url}
\usepackage[hidelinks]{hyperref}
\usepackage[utf8]{inputenc}
\usepackage[small]{caption}
\usepackage{graphicx}
\usepackage{amsmath}
\usepackage{booktabs}
\usepackage{algorithm}
\usepackage{algorithmic}
\usepackage[switch]{lineno}
\usepackage{xcolor}
\usepackage{siunitx}
\usepackage{multirow}
\usepackage{amsthm,thmtools, thm-restate}

\theoremstyle{definition}
\newtheorem{definition}{Definition}
\DeclareMathOperator{\IC}{IC}
\DeclareMathOperator{\degree}{deg}
\newcommand{\G}{\mathcal{G}}

\title{Rewarding Explainability in Drug Repurposing with Knowledge Graphs}

\author{
Susana Nunes$^{1,2}$
\and
Samy Badreddine$^{2,3,4}$\and
Catia Pesquita$^1$\\
\affiliations
$^1$LASIGE, Faculty of Sciences, University of Lisbon, Lisbon, Portugal\\
$^2$Sony AI, Barcelona, Spain\\
$^3$University of Trento, Trento, Italy\\
$^4$Bruno Kessler Institute, Trento, Italy
\emails
scnunes@ciencias.ulisboa.pt
}

\begin{document}
\maketitle

\begin{abstract}
Knowledge graphs (KGs) are powerful tools for modelling complex, multi-relational data and supporting hypothesis generation, particularly in applications like drug repurposing. However, for predictive methods to gain acceptance as credible scientific tools, they must ensure not only accuracy but also the capacity to offer meaningful scientific explanations.

This paper presents a novel approach REx, for generating scientific explanations based in link prediction in knowledge graphs. It employs reward and policy mechanisms that consider desirable properties of scientific explanation to guide a reinforcement learning agent in the identification of explanatory paths within a KG. The approach further enriches explanatory paths with domain-specific ontologies, ensuring that the explanations are both insightful and grounded in established biomedical knowledge. 

We evaluate our approach in drug repurposing using three popular knowledge graph benchmarks. The results clearly demonstrate its ability to generate explanations that validate predictive insights against biomedical knowledge and that outperform the state-of-the-art approaches in predictive performance, establishing REx as a relevant contribution to advance AI-driven scientific discovery.
\end{abstract}

\section{Introduction}
Knowledge Graphs (KGs) have emerged as versatile representations for capturing complex, multi-relational data in various scientific domains. They play a critical role in organizing, exploring, and sharing knowledge while supporting AI-based scientific discovery by providing structured, conceptually rich frameworks that can align neural model predictions with domain knowledge\cite{d2024role}. Link prediction has proven to be a powerful tool for hypothesis generation, enabling the discovery of novel relations between entities \cite{ott2022linkexplorer,akujuobi2024link}. Applications include gene-disease associations~\cite{yuen2020better} and drug repurposing~\cite{napolitano2018gene2drug}, where new therapeutic targets are discovered for existing drugs. 

To serve effectively as a scientific tool, artificial intelligence must possess the capability to generate \textit{scientific explanations}~\cite{duran2021dissecting}. This raises an important question: is there a fundamental relationship between the explanation of natural phenomena and the explanation of algorithmic outputs? This topic is subject to ongoing debate among researchers, particularly regarding how to epistemically ground the results of computational models as reliable representations of real-world phenomena. Nonetheless, there is widespread consensus that computational artefacts can facilitate the explanation of natural phenomena~\cite{duran2017varying,krohs2008digital,duran2021dissecting}.

However, recent studies highlight that widely used attribution models \cite{ribeiro2016should,lundberg2017unified} are inadequate for achieving the level of scientific and human-level explainability required for meaningful insights~\cite{chou2022counterfactuals}. This is also true of state-of-the-art link prediction explainability approaches~\cite{rossi2022kelpie,betz2022adversarial}, which are limited to identifying relevant features or triples, falling short of fulfilling the requirements of scientific explanations. In fact, relevant theories of scientific explanation argue that purely statistical explanations fail to identify causally relevant factors or mechanisms~\cite{salmon1984scientific}.

For these predictive methods to be adopted as reliable scientific tools, they must not only deliver accurate results but also afford mechanisms for \textit{scientific explanations}, i.e., methods that explain the scientific validity of the predictions, ensuring that they make sense with the current scientific body of knowledge and are not the result of spurious correlations~\cite{holzinger2019causability}. Take as an example the following explanation for a drug recommendation for the patient John Doe that is grounded on a specific inhibitory mechanism that mitigates the effects of a deleterious mutation:
   \textit{ \textit{John Doe} --has mutation$\rightarrow$ \textit{MET T540G} --part of$\rightarrow$  \textit{MET Gene} --related to $\rightarrow$  \textit{Tyrosine Kinase Activity} -- inhibited by $\rightarrow$ }\textit{Sunitinib}.

Moreover, the potential of KGs to support scientific insights extends beyond link prediction. As long as inputs and outputs of a hypothesis generation system can be represented within a KG, the scientific knowledge encoded therein can be explored to create scientific explanations. 

This paper focuses on generating knowledge-driven scientific explanations to validate scientific hypotheses generated by AI methods. More specifically, we focus on the identification of explanatory KG paths in drug repurposing. This task can be framed as a link prediction problem, where the goal is to generate scientific hypotheses by inferring potential new relationships between known entities, e.g., \textit{(minoxidil, treats, hair loss)}.

Our novel approach employs reward and policy mechanisms that consider desirable properties of scientific explanation to guide a reinforcement learning (RL) agent in the identification of the explanatory paths. The approach further enriches explanatory paths with domain-specific ontologies, ensuring that the explanations are both insightful and grounded in established biomedical knowledge.

Our contributions include: (1) a method to extend RL frameworks to consider scientific explainability properties when generating explanatory paths; (2) a method to calculate the relevance of explanatory paths that accounts for research bias; (3) a method to compose fully-fledged scientific explanations by integrating relevant paths with descriptive ontology classes; (4) the first evaluation of knowledge-driven drug repurposing explanation on three distinct benchmark KGs.

\section{Related Work}
KGs have emerged as critical tools in Explainable AI due to their ability to model multi-relational data and generate interpretable explanations. Despite significant advances, existing methods face several challenges, particularly in domains like biomedicine, where biologically relevant explanations are essential for scientific validation.

In the biomedical domain, KGs have been applied to justify AI-driven predictions for drug repurposing. For example, PoLo \cite{liu2021neural} combines representation learning with logical constraints to identify interpretable reasoning paths. Similarly, Ozkan et al. \cite{ozkan2023generating} extended the PREDICT framework~\cite{gottlieb2011predict} to rank explanatory paths by their relevance to drug indications, incorporating established biomedical relationships. Stork et al. \cite{stork2023explainable} proposed to improve RL-based drug repurposing using phenotype annotations. These efforts build on RL frameworks for multi-hop inference, such as DeepPath \cite{xiong2017deeppath} and MINERVA~\cite{das2017minerva}. However, they prioritize predictive accuracy over other desirable properties for scientific explanation, such as relevance, limiting their utility in scientific contexts.

\section{Problem Definition}
Consider a knowledge graph $\G = (\mathcal{E}, \mathcal{R}, F)$, where $\mathcal{E}$ is a set of entities, $\mathcal{R}$ is a set of relations, and $F \subseteq \mathcal{E} \times \mathcal{R} \times \mathcal{E}$ is a set of triples denoted as $(s,r,o)$ for subject, relation, and object. We will often abuse notation and directly define $\G$ as the set of triples such that we can write $(s,r,o) \in \G$.

Modern theories of scientific explanation often emphasize explanatory virtues such as empirical adequacy, simplicity, scope, and coherence. We adopt the taxonomy proposed in~\cite{keas2018systematizing} for scientific theories where four types of explanatory virtues are identified: (i) Evidential theoretical virtues: evidential accuracy, causal adequacy, and explanatory depth; (ii) Coherential theoretical virtues: internal consistency, internal coherence, and universal coherence; (iii) Aesthetic theoretical virtues: beauty, simplicity, and unification; (iv) Diachronic theoretical virtues: durability, fruitfulness, and applicability.

By virtue of being domain models and representing explicit relations between entities, KGs can naturally support causal adequacy and contribute to explanatory depth by identifying causal mechanisms, e.g., \textit{vincristine --treats$\rightarrow$ lymphatic system cancer --associated with$\rightarrow$ TIA 1 gene $\leftarrow$ associated with-- hematologic cancer.} Path and rule extraction are common methods to offer explanations in graph theory and link prediction \cite{meilicke2019anytime,liu2021neural,zhang2023page} that align well with evidential virtues, in particular causal adequacy. Moreover, KGs can also cover all coherential virtues by affording mechanisms to ensure internal consistency and coherence (i.e., an explanation's components are not contradictory) and universal coherence (i.e., the explanation fits well with extant knowledge). While aesthetic virtues are generally considered less valuable, both their pragmatic value --- simpler or shorter explanations are easier to grasp --- and epistemic value (v. Occam's Razor) are relevant. Finally, diachronic virtues require additional time after the initial explanation formulation and are therefore out of our scope.

Having established paths as the core of our explanation definition, we define a path of length $k$ in $\G$ as a finite sequence of triples $(e_i,r_i,e_{i+1}) \in \G$ for $i=1,\dots,k-1$, which joins a sequence of distinct entities $e_1, \dots, e_k \in \mathcal{E}$.
However, not all paths connecting the subject and object of a hypothesis triple fit the criteria of scientific explainability. 
In what regards to evidential virtues, a first challenge is ensuring \textit{causal detail}, which often translates to producing a chain of intermediaries linking the entities at hand~\cite{rosales2021scientific}. A second challenge is ensuring the \textit{relevance} of explanations since an explanation that achieves causal detail can still be vague and afford little scientific insight. For example, an explanation for the hypothesis \textit{(sunitinib, treats, renal cancer)} that takes the form \textit{(sunitinib, is a, antineoplastic agent, treats, cancer, super class of, renal cancer)} is correct, causal, but not scientifically relevant. A third challenge lies in ensuring the \textit{completeness} of explanations since an adequate causal account often requires the interaction of multiple factors rather than a single directed cause-and-effect path.
Ensuring \textit{universal coherence} can also be a challenge since many popular scientific KGs do not possess a schema backed by an ontology, which limits the ability to ensure and evaluate the logical coherence with the domain. Finally, a fifth challenge is related to \textit{simplicity} or parsimony and how to ensure the pragmaticity of explanations without sacrificing other relevant properties. Clearly, addressing the first four challenges necessarily represents a trade-off with addressing the fifth since ensuring a detailed, complete, relevant and universally coherent explanation very likely requires a larger and more complex explanation.

\section{Method}
Our goal is to search, given a source node $s$, for paths to target nodes $o$ seen in training and to \emph{generalize} paths to unseen targets during inference. To the best of our knowledge, no exact path-finding algorithm fits our problem, as they typically require fixed source and target nodes and lack inherent generalization. While heuristic search might offer some generalization, we opted for RL due to its scalability, flexibility, and crucial generalization capabilities.
\subsection{Overview }

Our approach generates scientific explanations for a hypothesis \textit{h} --- described as triple in a KG $\mathcal{G}$ --- as a subgraph $\mathcal{G}_h$ that integrates a set of relevant explanatory paths $p \in P$ and relevant ontology classes that describe entities in the path. An explanatory path for hypothesis \textit{h} is a path that connects the subject and object entities of \textit{h}, respectively $s_h$ and $o_h$, through a chain of relevant related entities.

Our explanation generation strategy addresses the challenges of causal detail and relevance by using reinforcement learning, conditioned on the hypothesis to validate, to find explanatory paths. It employs a reward-shaping mechanism to ensure multi-objective optimization regarding fidelity (i.e., to ensure paths successfully connect $s_h$ and $o_h$) and relevance (aiming to maximize the information content (IC) of a path, a measure of the specificity of the entities composing it). Simplicity is ensured by a policy that ensures the RL agent produces paths without loops and that terminate when $o_h$ is reached.
These explanatory paths are then filtered to include those that are maximally relevant and representative of different explanation types, ensuring \textit{completeness}. These are grouped to form the backbone of $\mathcal{G}_h$, which is then enriched by including type axioms connecting entities in $\mathcal{G}_h$ to relevant ontology classes, thereby affording a richer contextualization of the explanation subgraph, facilitating \textit{universal coherence} 

\begin{figure*}[ht]
    \centering
    \includegraphics[width=1.0\textwidth, trim=40 40 10 30, clip]{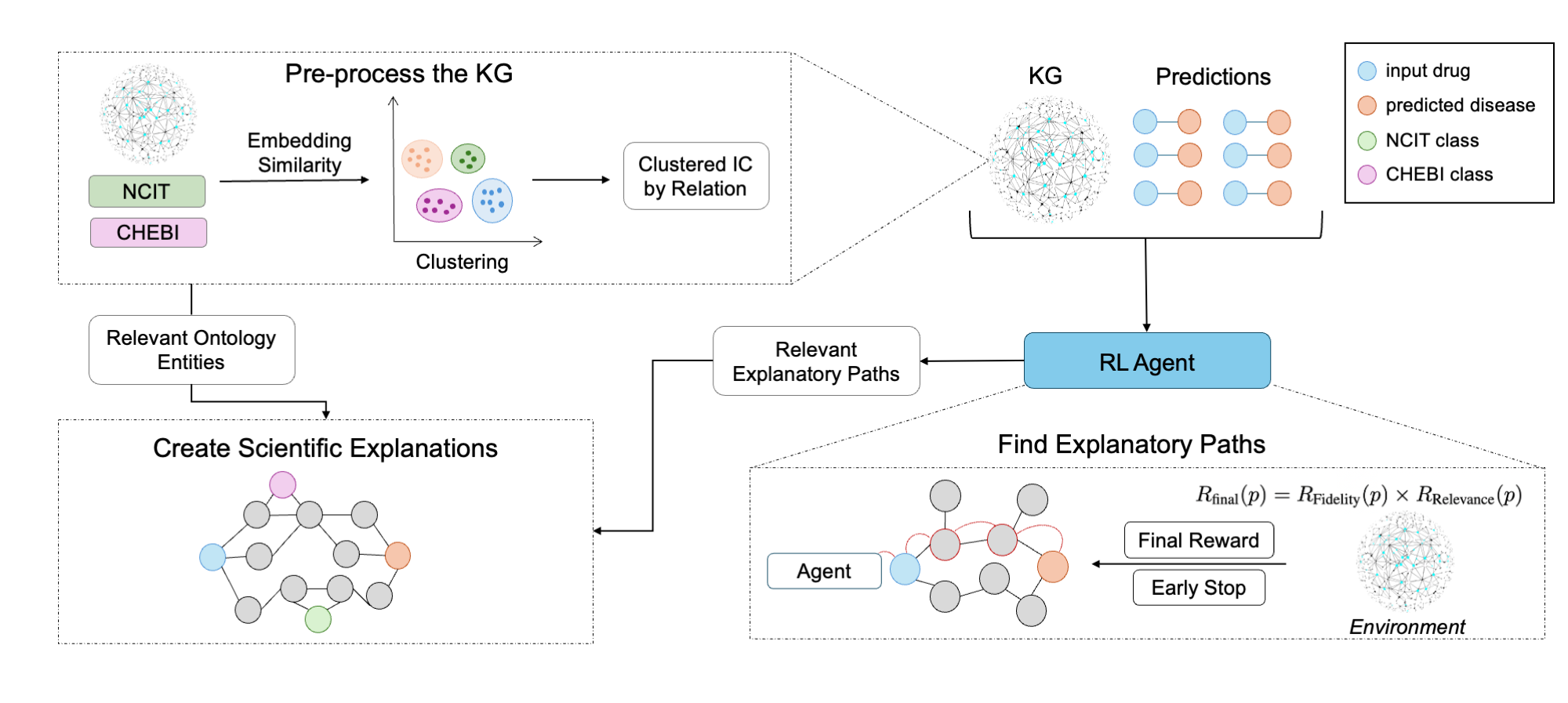}   
    \caption{Overview of the approach to generate scientific explanations, with three main phases: (1) pre-process KG, where the information content is computed, (2) find explanatory paths, where a reinforcement learning agent is trained, and (3) create scientific explanations, where the learned paths are integrated with relevant ontologies to improve context.}
    \label{fig:overview}
\end{figure*}

The overall approach is illustrated in Figure \ref{fig:overview}. Given a biomedical KG and a set of drug repurposing predictions to be individually explained, the method follows a three-phase process: (1) computing the information content of entities; (2) finding explanatory paths; (3) generating scientific explanations.

\subsection{Information Content}
An essential aspect of our method is to compute the relevance of paths. Our hypothesis is that paths involving less frequent entities are more likely to reveal meaningful relationships, resulting in explanations that are both insightful and representative of the underlying scientific knowledge. We define the relevance of a path as the average of the information content (IC) of the edges that compose it.

We formalize the concept of IC from information theory, taking into account \textit{node degree counts}: the number of edges (relations) connected to a node (entity). First, we define the IC of an entity $v$ that appears in a triple, either as the subject or object. The informativeness of an edge is determined by the average IC of the two entities it connects. To compute the IC of an entity, we introduce the concept of \textit{Clustered IC}, which refines the IC calculation by analyzing a clustered graph instead of the original graph. The intuition is to diminish possible bias resulting from heterogeneous granularity levels due to over-studied and under-studied areas, as well as cases where minor variations of the same concept are included. This can be further refined by evaluating nodes within the context of specific relations, reflecting the intuition that the significance of an entity may vary depending on the nature of the relationships it engages in, providing a more detailed and accurate measure of path informativeness.

\subsubsection{IC of a Node}
Let $T = (S,R,O)$ be a random variable for the KG triples. Sampling from $T$ means sampling a random triple from the graph. 

\begin{definition}
   The IC of a node $v \in \mathcal{E}$, denoted $\IC(v)$, is defined by the information content of the event $(S=v) \cup (O=v)$. It measures the surprisal of a node appearing as a subject or object in a randomly sampled triple.
\end{definition}

We can derive the following (proof in appendix):
\begin{restatable}{theorem}{nodeIC}
    \label{thm:nodeIC}
    Considering each triple of the graph as independently and identically distributed, we have
    \begin{equation}
        \IC(v) = -\log \frac{\degree(v)}{|\G|}.
    \end{equation}
    where $\degree(v)$ the total degree of an entity node in the knowledge graph,
\end{restatable}

\subsubsection{Clustered IC of a Node}
The IC of a node can be modified to account for potential node degree bias due to different granularities, whereby in some subdomains, two very similar concepts are represented by different entities. This can be an effect of research bias~\cite{reynolds2021accelerating} and not necessarily translate to a scientifically meaningful measure of frequency.
Let $A$ and $A'$ be two such similar entities and $B$ a third entity. 
Two possible issues can arise: (i) non-meaningful relations between $A$ and $A'$ (\emph{is a} or \emph{synonym of}) can "boost" their node degrees, (ii) inversely, some relations highlighted only between $A'$ and $B$ but not between $A$ and $B$ (whereas, in practice, they should hold) can "hide" the true node degree of $A$. 
Such issues can artificially increase or decrease the node degree of concepts.
The clustered graph aims to group semantically similar concepts together to mitigate this.

Consider the set of clusters $\mathcal{C} = \{C_{1}, C_{2}, \dots, C_{k}\}$ such that each $C_{i} \subseteq \mathcal{E}$ and $\bigcup_{i=1}^k C_i=\mathcal{E}$ . Consider the clustering function $\kappa: \mathcal{E} \rightarrow \mathcal{C}$. Each entity $v \in \mathcal{E}$ belongs to exactly one cluster $\kappa(v) = C_i$. The clusters are assumed to group entities per semantic similarity.

Given the KG $\G=(\mathcal{C}, \mathcal{R}, F)$, we derive a clustered graph $\G_c=(\mathcal{C}, \mathcal{R}, F_c)$ where $F_{c}= \{(C_{i},r,C_{j}) | \exists (u,r,v) \in F \text{ s.t. } u \in C_{i} \text{ and } v \in C_j \}$. Intuitively, the clustered graph derived from $\G$ is a graph where the nodes are grouped into clusters and there is an edge between two clusters if any nodes in these clusters were connected by an edge in $\G$.

As for the original graph, we derive random variables for the clustered graph $(S_{c}, R_{c}, O_{c})$ .
\begin{definition}
    The clustered IC (CIC) of a node $v \in \mathcal{E}$ is defined by the information content of the event $(S_{c}= \kappa(v)) \cup (O_{c} = \kappa(v))$. It measures the surprisal of a node belonging to one of the subject cluster or object cluster randomly sampled in a triple from the clustered graph $\G_c$.  
\end{definition}

Similar to before, we can derive (proof in appendix):
\begin{restatable}{theorem}{clusterednodeIC}
    \label{thm:clusterednodeIC}
    Considering each triple of the clustered graph as independently and identically distributed, we have
    \begin{equation}
        \IC_c(v) = -\log \frac{\degree(\kappa(v))}{|\G_c|}
    \end{equation}
    where the degree of $\kappa(v)$ is calculated in the clustered graph.
\end{restatable}

\begin{definition}
The Clustered IC of a node by relation type is a variant of CIC where node degree considers only edges of a given type $\degree(v,r)$.
\end{definition} 
To support the CIC computation, we generated embeddings using OWL2vec*~\cite{chen2021owl2vec}, which were then clustered using K-means, with the number of clusters set to 10\% of the total node count. 

\subsection{Finding Explanatory Paths}
We adapt the RL path-finding strategy proposed in~\cite{das2017minerva} to our hypothesis validation purpose. It specifies a deterministic partially observed Markov decision process as a 5-tuple (S, O, A, T, R).

\textbf{States.} The state space $\mathcal{S}$ consists of all combinations in \(\mathcal{E} \times \mathcal{E} \times \mathcal{E}\). Intuitively, we want a state to encode the hypothesis subject $s_h$ and object $o_h$, as well as a location of exploration $e$ (current location of the RL agent). Therefore, a state $S \in \mathcal{S}$ is represented by:
\[
S = (e, s_h, o_h)
\]

\textbf{Observations.} The complete state of the environment is not observed since the agent only knows its current location $e$ and the hypothesis subject. Formally, the observation function $O : S \to \mathcal{E}  \times \mathcal{E}$ is defined as:
\[
O = (e, s_h)
\]

\textbf{Actions.} The set of possible actions \(A_S\) from a state \(S = (e, s_h, o_h)\) include all edges connected to the current node \(e\) in \(G\) or a decision to stop. Formally:
\[
A_S = \{(e, r, e_d) \in F : S = (e, s_h, o_h), r \in R, e \in \mathcal{E} \} \cup \{\text{STOP}\}
\]

This means that at each state, the agent decides either to stop or to continue to destination node $e_d$.

\textbf{Transition.} Environment evolution is deterministic, simply updating the state to the new entity selected by the agent.

\textbf{Rewards.} The reward function captures two objectives: 
\begin{itemize}

\item \textbf{Fidelity}: A scientific explanation should necessarily align with the hypothesis. Fidelity indicates whether the path-finding algorithm successfully connects $s_h$ and $o_h$. Formally, if $S_T = (e, s_h, o_h)$ is the end state and $e=o_h$, $R_{\text{Fidelity}(p)}=1$, else $R_{\text{Fidelity}(p)}=0$.

\item \textbf{Relevance}: A scientific explanation should provide detailed insights into the mechanisms underlying the hypothesis. Formally, when the end state is reached, the average IC of the path \textit{p} is computed to arrive at $R_{\text{Relevance}}(p)$. 
\end{itemize}

Formally, the final reward \( R_{\text{final}} \) of a path \textit{p} is given by:  

\[
R_{\text{final}}(p) =  R_{\text{Fidelity}}(p) \times R_{\text{Relevance}}(p)
\]  

\textbf{Policy Network.} We extend the policy proposed in~\cite{das2017minerva} with an early stopping mechanism. This policy --- based on LSTMs to encode the history of actions and observations --- presents desirable properties for explanatory path generation on KGs, namely that it is permutation-invariant to edge ordering and history-dependent, with decisions \(d_t\) mapping the history \(H_t\) to a probability distribution over available actions \(A_{S_t}\). 
The history \(H_t = (H_{t-1}, A_{t-1}, O_t)\) records past actions (\(A_{t-1}\)) and observations (\(O_t\)). When the policy network chooses an action from all available actions, $\mathcal{A}_{S_t}$
 if $S_T = (e, s_h, o_h)$ and $e=o_h$, the agent does not take any further actions. This mechanism not only reduces unnecessary exploration but also promotes simplicity. The original policy resorted to a special action which goes from a node to itself, which resulted in possible loops and repetitiveness.

%\vspace{1cm}
\textbf{Training.} We extend training to 30 rollouts, following~\cite{liu2021neural}.

\subsection{Generating Scientific Explanations}
To construct scientific explanations for predictions, we begin by analyzing all explanatory paths found in the previous step and grouping them based on their metapaths, i.e., the sequence of entity types and relations in the path. For each metapath, we select the path with the highest IC. These selected paths are then merged into a graph and enriched with the lowest common ancestors (LCA) between all consecutive entities in a path. More formally,
\[\mathcal{G}_\text{h} = \bigcup_{p \in \mathcal{P}} p  \quad \cup \bigcup_{(e_i, e_{i+1}) \in p} \text{LCA}(e_i, e_{i+1})
\]

\section{Experiments}
\subsection{Drug repurposing}
To evaluate the effectiveness of REx\footnote{Code and Supplementary Material available at https://github.com/liseda-lab/REx.} in generating scientifically valid explanations, we applied it to the task of validating drug repurposing hypotheses. Drug repurposing identifies new therapeutic uses for existing drugs, and such a hypothesis can be formulated as a triple $(drug, treats, disease)$. As benchmarks for our experiments, we used well-known biomedical KGs that describe drugs, diseases and other relevant entities for drug repurposing\footnote{Data repository links in Supplementary Material.}: Hetionet~\cite{himmelstein2017systematic}, PrimeKG~\cite{chandak2023building}, and OREGANO~\cite{boudin2023oregano}. Hetionet is an integrative biomedical KG combining data from 29 sources, including genes, compounds, and diseases, and with more than 45,000 entities. PrimeKG is a precision medicine-oriented knowledge graph spanning multiple biological scales, such as pathways, phenotypes, and drug indications, with nearly 130,000 entities. OREGANO is specifically designed for drug repurposing, aligning experimental data with drug-disease associations and more than 98,000 entities. In each case, inverse edges were added when not provided in the KG. More detailed statistics can be found in Supplementary Material. 

\subsection{KG Enrichment}
Since the benchmark KGs are semantically shallow and do not include an ontology-based schema, we enriched each KG by aligning it to relevant domain ontologies, specifically the National Cancer Institute Thesaurus (NCIT)~\cite{hartel2005modeling} and the Chemical Entities of Biological Interest (ChEBI)~\cite{degtyarenko2007chebi}, which accurately describe drugs, diseases and other entities in the KGs. These alignments were generated using the ontology matching system AML~\cite{faria2023agreementmakerlight}.

To evaluate explanation relevance, the paths identified in Hetionet were transformed into metapaths. These metapaths were then compared to a ground truth derived from the findings of Himmelstein~\cite{himmelstein2017systematic}.

\section{Results and Discussion}

\subsection{Predictive Performance Evaluation}
 We evaluated the predictive performance of REx’s explanatory paths against several baseline methods, including MINERVA, a RL-based method that answers queries through multi-hop reasoning~\cite{das2017minerva} and PoLo, that extends it with logical constraints to improve interpretability~\cite{liu2021neural} (details in Supp. Material).

This evaluation assesses the most basic property of a scientific explanation --- if it aligns with the hypothesis. The reported values for REx, PoLo, and MINERVA correspond to the calculation of a mean across five independent successful training runs, with a standard deviation between 0.004 and 0.028. All other results were reported in~\cite{liu2021neural}.

Table \ref{tab:results_hetionet} presents the results of various methods on the Hetionet KG. REx outperformed all state-of-the-art methods, achieving the highest MRR of 0.427. This indicates that REx effectively integrates fidelity, simplicity, and relevance to find robust explanations.

\begin{table}[h!]
\centering
\begin{tabular}{lcccc}
\toprule
\textbf{Method} & \textbf{Hits@1} & \textbf{Hits@3} & \textbf{Hits@10} & \textbf{MRR} \\ \midrule
AnyBURL         & 0.229           & 0.375           & 0.553            & 0.322        \\ 
TransE          & 0.099           & 0.199           & 0.444            & 0.205        \\ 
DistMult        & 0.185           & 0.305           & 0.510            & 0.287        \\ 
ComplEx         & 0.152           & 0.285           & 0.470            & 0.250        \\ 
ConvE           & 0.100           & 0.225           & 0.318            & 0.180        \\ 
RESCAL          & 0.106           & 0.166           & 0.377            & 0.187        \\ 
R-GCN           & 0.026           & 0.245           & 0.272            & 0.135        \\ 
CompGCN         & 0.172           & 0.318           & 0.543            & 0.292        \\ 
pLogicNet       & 0.225           & 0.364           & 0.523            & 0.333        \\ \midrule 
MINERVA         & 0.264           & 0.409           & 0.593            & 0.370        \\ 
PoLo            & 0.314           & 0.428           & \textbf{0.609}            & 0.402        \\ 
REx & \textbf{0.338} & \textbf{0.461} & \textbf{0.609} & \textbf{0.427} \\

\bottomrule
\end{tabular}
\caption{Performance comparison of various methods for predictions on Hetionet based on Hits@k and MRR metrics.}
\label{tab:results_hetionet}
\end{table}

Similarly, Table \ref{tab:results_primekg} and Table \ref{tab:results_oregano} highlight the results on PrimeKG and OREGANO with the best-performing methods on Hetionet. Here, REx once again surpassed the state of the art, achieving an MRR of 0.376 on PrimeKG and 0.278 on OREGANO. These results confirm that REx is able to produce explanations that have more predictive power than comparable methods across a variety of KGs.

\begin{table}[h!]
\centering
\begin{tabular}{lcccc}
\toprule
\textbf{Method} & \textbf{Hits@1} & \textbf{Hits@3} & \textbf{Hits@10} & \textbf{MRR} \\ \midrule
MINERVA         & 0.262            & 0.420             & \textbf{0.546 }             & 0.359      \\ 
PoLo            & 0.245             & 0.408             & 0.526              & 0.344        \\ 
REx & \textbf{0.286}        & \textbf{0.429 }         & 0.544          & \textbf{0.376} \\\bottomrule
\end{tabular}
\caption{Performance comparison for predictions on PrimeKG based on Hits@k and MRR metrics.}
\label{tab:results_primekg}
\end{table}

\begin{table}[h!]
\centering
\begin{tabular}{lcccc}
\toprule
\textbf{Method} & \textbf{Hits@1} & \textbf{Hits@3} & \textbf{Hits@10} & \textbf{MRR} \\ \midrule
MINERVA         & 0.133           & 0.200           & 0.489             & 0.220     \\ 
PoLo            & \textbf{0.171}            & 0.292            & 0.473              & 0.259        \\ 
REx &  \textbf{0.171} & \textbf{0.327}          & \textbf{0.533}        & \textbf{0.278} \\\bottomrule
\end{tabular}
\caption{Performance comparison for predictions on OREGANO based on Hits@k and MRR metrics.}
\label{tab:results_oregano}
\end{table}

\begin{figure}[ht]
    \centering
\includegraphics[width=\columnwidth]{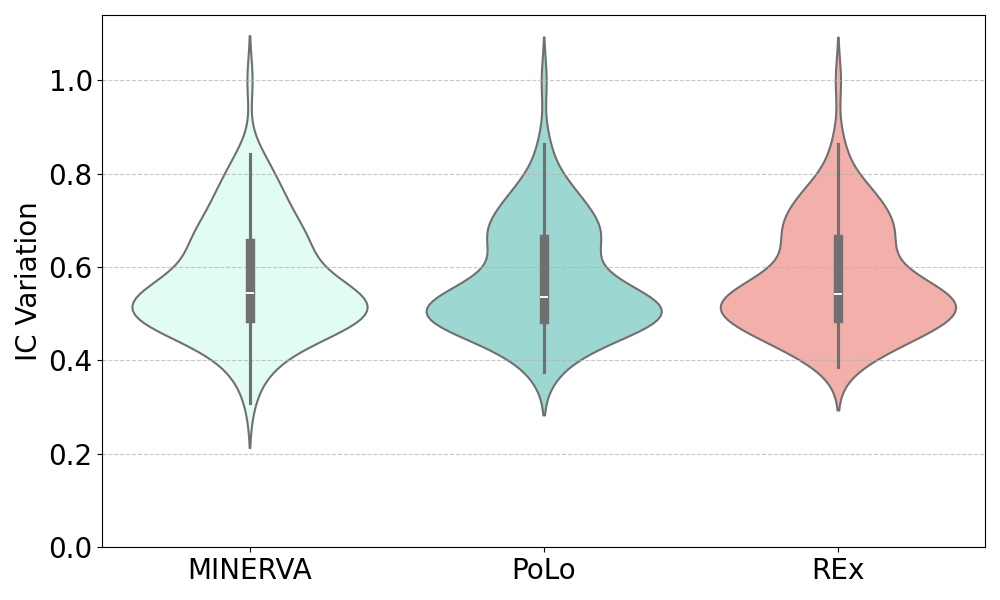}
    \caption{IC distribution for the methods MINERVA, PoLo, and REx using the Hetionet dataset.}
    \label{fig:ic_distribution}
\end{figure}

To evaluate the relevance of explanatory paths, we computed the average IC for each path generated with REx, MINERVA, and PoLo. The distribution of ICs shown in Figure~\ref{fig:ic_distribution} clearly indicates that REx's paths have a higher IC in general and that it does not generate paths of low relevance, with the vast majority above 0.4. While PoLo does not fall far behind, MINERVA produces a larger portion of paths with lower IC. This underlines that the reward mechanism employed by REx effectively excludes low-relevance paths.

To analyze the contribution of each component to the overall performance of REx, we conducted an ablation study by systematically removing individual components of the approach:
\begin{itemize}
    \item \textbf{REx}: The complete version of our approach. 
    \item \textbf{REx \textit{-s}}: A variation where the early stop mechanism is removed, sacrificing the simplicity virtue.
    \item \textbf{REx \textit{-r}}: A variation where the relevance is excluded from the final reward calculation.
    \item \textbf{REx \textit{-rs}}: A variation where both simplicity and relevance are not considered, leaving only fidelity as a final reward. 
\end{itemize}

The results shown in Table \ref{tab:results_ablation} indicate that when the early stopping mechanism was removed (-s), the MRR dropped in all datasets. This result highlights the importance of generating concise explanations, as longer paths can increase complexity and hinder interpretability. Similarly, the removal of relevance (-r) resulted in a noticeable decrease in performance, especially for Hetionet. This demonstrates that incorporating information content to ensure biologically meaningful paths is crucial for finding scientifically valid explanations. The combination of removing both the early stopping mechanism and relevance (-rs) led to the most significant reduction in performance in the cases of PrimeKG and OREGANO, emphasizing the complementary roles of these two components.

\begin{table}[h!]
\centering
\begin{tabular}{@{}llcccc@{}}
\toprule
\textbf{KG} & \textbf{Method} & \textbf{Hits@1} & \textbf{Hits@3} & \textbf{Hits@10} & \textbf{MRR} \\ \midrule
\multirow{4}{*}{\rotatebox[origin=c]{90}{Hetionet}} 
& REx \textit{-s}  & \textit{0.309} & \textit{0.446}          & \textbf{0.627}          & \textit{0.407} \\
& REx \textit{-r}  & 0.295          & 0.432 & 0.600 & 0.392          \\
& REx \textit{-rs} & 0.302          & \textit{0.446 }         & \textit{0.609}          & 0.404          \\
& REx              & \textbf{0.338} & \textbf{0.461} & \textit{0.609} & \textbf{0.427} \\ \midrule
\multirow{4}{*}{\rotatebox[origin=c]{90}{PrimeKG}} 
& REx \textit{-s}  & 0.278          & 0.426          & \textit{0.544}          & 0.370          \\
& REx \textit{-r}  & \textit{0.284}  & \textbf{0.431}  & \textbf{0.554}   & \textbf{0.376}         \\
& REx \textit{-rs} & 0.277          & \textit{0.429 }         & 0.540          & 0.369         \\
& REx              & \textbf{0.286}        & \textit{0.429 }         & \textit{0.544}          & \textbf{0.376}  \\ \midrule
\multirow{4}{*}{\rotatebox[origin=c]{90}{OREGANO}} 
& REx \textit{-s}  & 0.143          & 0.314          & 0.543          & 0.264          \\
& REx \textit{-r}  & 0.149          & 0.244          & 0.514          & 0.244          \\
& REx \textit{-rs} & 0.105          & 0.222          & 0.498          & 0.209          \\
& REx              & \textbf{0.171} & \textbf{0.327}          & \textbf{0.533}          & \textbf{0.278}          \\ \\
\bottomrule
\end{tabular}
\caption{Ablation of REx based on hits@k and MRR. Bold indicates best result, italics second best.}
\label{tab:results_ablation}
\end{table}

Table~\ref{tab:results_ic} presents a comparative evaluation of different approaches for computing IC within the REx using Hetionet. 
The results show that Clustered IC by Relation yields the highest scores overall, underscoring the value of tailoring IC calculations to specific relations. Interestingly, IC outperforms the Clustered IC method in every metric, indicating that an IC that is clustered and blind to relation type loses relevant information (details in Supp. Material). 

%TABLE OVERFLOW RESOLVED
\begin{table}[h!]
\centering
  \resizebox{\columnwidth}{!}{%
\begin{tabular}{lcccc}
\toprule
\textbf{Method} & \textbf{Hits@1} & \textbf{Hits@3} & \textbf{Hits@10} & \textbf{MRR} \\ \midrule
IC        & 0.290           & 0.437           & 0.595             & 0.391     \\ 
CIC            & 0.264            & 0.419            & 0.591              & 0.370        \\ 
CIC by Relation &  \textbf{0.338} & \textbf{0.461} & \textbf{0.609} & \textbf{0.427} \\\bottomrule
\end{tabular}
}
\caption{Performance comparison for different types of IC on REx using Hetionet based on hits@k and MRR.}
\label{tab:results_ic}
\end{table}

\subsection{Ground-Truth Evaluation}
To assess the relevance of the explanatory paths generated by REx, we compared them to the ground truth paths identified in \cite{himmelstein2017systematic}, which are recognized as key mechanisms for drug repurposing. This comparative approach draws on the principle of analogy \cite{thagard1978best,thagard1989explanatory}, where new explanatory mechanisms gain credibility when they align with well-established causal structures. 

Our analysis showed that REx identified 12 distinct types of explanatory paths. Of these, 8 path types were fully consistent with the ground truth, confirming their biological plausibility and alignment with existing biomedical knowledge. The remaining 5 path types (in Supp. Material), while not explicitly included in the ground truth, still provided biologically coherent insights that could help formulate novel hypotheses about drug–disease relationships. In fact, the majority of these paths are similar to ground truth ones but use \textit{palliates} instead of \textit{treats}, which is semantically similar. By uncovering both validated and new explanatory paths, REx demonstrates its ability to both replicate known mechanisms and offer novel and plausible explanations that may advance our understanding of drug repurposing. 

Regarding path frequency in the datasets, each knowledge graph exhibits distinct mechanistic preferences (in Supp. Material), with Hetionet showing a clear preference for side effect-based paths, Oregano emphasizing gene-mediated paths and PrimeKG generating a substantial number of paths solely based on (\textit{drug -- indication --- disease}) chains. The distribution of path frequencies is similarly skewed across all KGs, with a few highly frequent paths and many rare ones. This suggests that while certain drug repurposing mechanisms are well-described, there may be numerous specialized pathways that might be relevant for specific cases. Further analysis revealed that multiple explanatory paths can be identified for a single prediction, with an average of 12 relevant paths per drug repurposing hypothesis in Hetionet. This diversity of explanatory paths highlights the complexity of drug repurposing and the importance of considering multiple explanatory paths when producing an explanation.

\subsection{Domain Expert Evaluation}
We recruited two life sciences graduates familiar with drug repurposing to evaluate the validity of 10 randomly selected full REx-generated explanations for the Hetionet dataset %(Figure~\ref{fig:user_paths} in  
(in Supp. Material). Each REx explanation was presented alongside a corresponding MINERVA explanation.

\begin{figure}[ht]
    \centering
\includegraphics[width=\columnwidth]{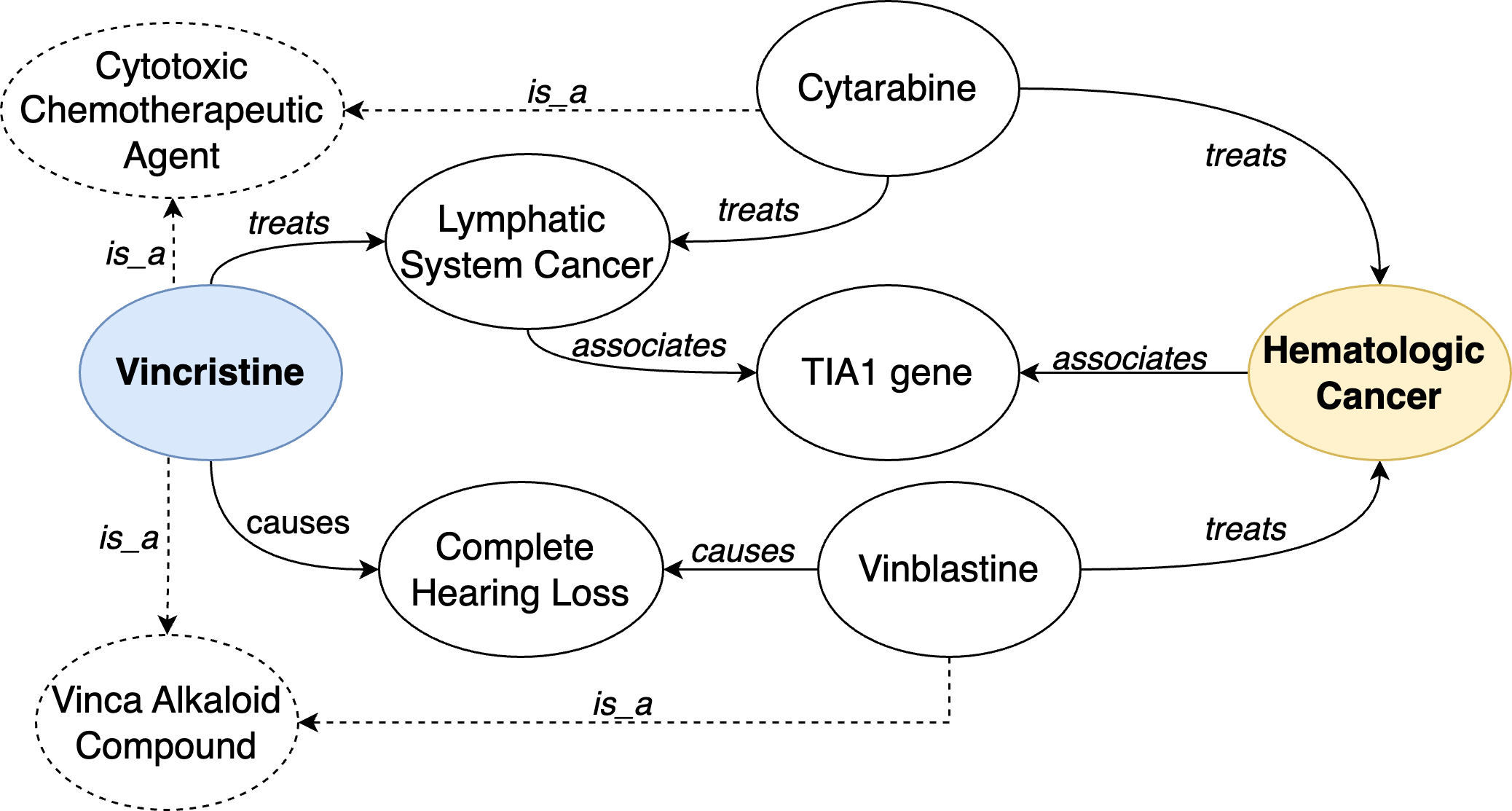}
    \caption{Explanation between Vincristine and Hematologic Cancer from Hetionet dataset.}
    \label{fig:case_study}
\end{figure}

\begin{figure}[ht]
    \centering
\includegraphics[width=\columnwidth]{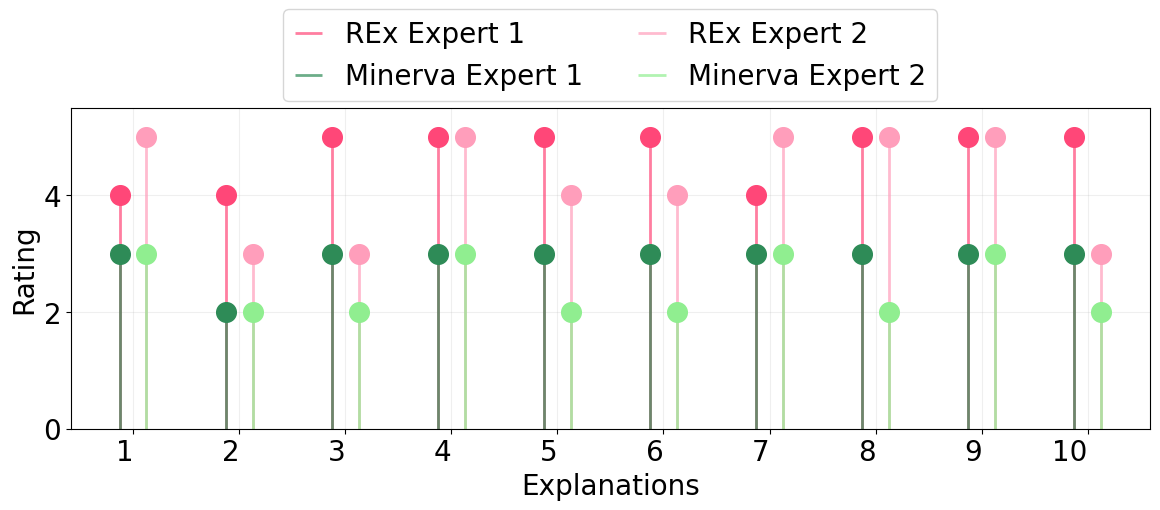}
    \caption{Comparison of the experts rating for 10 drug–disease repurposing explanations for Hetionet generated by REx (pink) and MINERVA (green).}
    \label{fig:user_study}
\end{figure}

The experts rated both explanations for each drug repurposing hypothesis on a scale from 1 to 5, reflecting their satisfaction with the explanation quality ranging from 1 (very low) to 5 (very high) and consulting any external references they deemed necessary. Figure~\ref{fig:user_study} depicts the experts' ratings for each explanation. Although both experts occasionally differ in how highly they rate the same method, the overall trend across all 10 explanations favours REx on every occasion. 

To complement this analysis, we provide a literature-based validation of a REx generated explanation for the hypothesis \textit{(vincristine, treats, hematologic cancer)} presented in Figure~\ref{fig:case_study}. A literature search revealed the scientific validity of this explanation, since both vincristine and vinblastine belong to the Vinca Alkaloid family, and both can cause neurotoxicity (including hearing loss)~\cite{madsen2019aspects}. Furthermore, cytarabine and vincristine are therapeutic options for lymphatic system cancers, which are associated with the TIA1 gene, which regulates the translation and stability of mRNAs involved in apoptosis, proliferation, and stress responses, relevant processes for cancer cell survival~\cite{sanchez2015t}. In turn, TIA1 is associated with hematologic cancers. 

\section{Conclusion}
We propose a novel method, REx, for generating scientific explanations of hypotheses based on KGs. Our method fulfils several desirable properties of scientific explanations, namely causal detail, relevance, completeness, coherence, and simplicity. It employs an RL-based approach guided by a dual reward that values both fidelity to the hypothesis and relevance to generate explanatory paths. The RL approach also considers an early stopping mechanism to consider simplicity. Paths are combined into a graph that is enriched with relevant ontology classes to ensure completeness and coherence. Notably, our approach can integrate a wide range of RL frameworks designed for graph-structured data, allowing our methodology to benefit from future evolutions in this field. 

REx outperforms the state of the art in predictive performance, produces more relevant explanatory paths and results in explanations that are considered of better quality by experts in three benchmark tasks for drug-repurposing hypothesis validation.
Nevertheless, the predictive performance of both REx and other state-of-the-art methods remains fairly modest, indicating that in many cases, no explanation can be found. This can be due to KG incompleteness or even to a lack of scientific evidence, so further analysis is required to elucidate this aspect. 

%Our experiments demonstrate that REx generalizes to different KGs, but generalization to other domains and applications is in principle possible. 
%Furthermore, REx can also be applied to hypotheses derived from other data sources or methods. These hypotheses can be modeled as new triples within a relevant KG, thereby extending the applicability of REx to non-KG-based hypothesis generation.

Our experiments show that REx generalizes across different KGs, and its extension to other domains and applications is, in principle, feasible. Furthermore, REx can be applied to hypotheses derived from alternative data sources or methods by modeling them as new triples within a relevant KG, thus broadening its applicability beyond KG-based hypothesis generation.

%Moreover, while the default maximum path length follows the widely accepted heuristic that humans can hold 7$\pm$2 items in short-term memory \cite{miller1956magical}, exploring the impact of longer or more complex paths on the scientific validity and usefulness of explanations remains an open question.

\section*{Acknowledgments}
This work was supported by FCT through the fellowship
2023.00653.BD, and the LASIGE Research Unit, ref. UID/00408/2025. It was also partially supported by the KATY project (European Union Horizon 2020 grant No. 101017453), and project 41, HfPT: Health from Portugal, funded by the Portuguese Plano de Recuperação e Resiliência. We thank Sony AI, where the first author conducted part of this work during an internship. We also thank Pedro Cotovio for his input on reinforcement learning fundamentals. 

\appendix

\bibliographystyle{named}
\bibliography{ijcai24}

\end{document}